# A Survey on Sensor Technologies for Unmanned Ground Vehicles

Qi Liu
School of Mechanical Engineering
Beijing Institute of Technology
Beijing, China
liuqibeishida@126.com

Shihua Yuan
School of Mechanical Engineering
Beijing Institute of Technology
Beijing, China
yuanshihua@bit.edu.cn

Zirui Li
School of Mechanical Engineering
Beijing Institute of Technology
Beijing, China
3120195255@bit.edu.cn

*Abstract*—Unmanned ground vehicles (UGVs) have a huge development potential in both civilian and military fields, and have become the focus of research in various countries. In addition, high-precision, high-reliability sensors are significant for UGVs' efficient operation. This paper proposes a brief review on sensor technologies for UGVs. Firstly, characteristics of various sensors are introduced. Then the strengths and weaknesses of different sensors as well as their application scenarios are compared. Furthermore, sensor applications in some existing UGVs are summarized. Finally, the hotspots of sensor technologies are forecasted to point the development direction.

*Keywords—unmanned ground vehicles, sensor, application*

## I. Introduction

The unmanned ground vehicle (UGV) is a comprehensive intelligent system that integrates functions including environmental perception, localization, navigation, path planning, decision-making and motion control [1]. It depends on various technologies such as sensors, computer science, and data fusion to operate efficiently [2].

For the civil application, UGV is mainly reflected in autonomous driving. The fusion of environmental perception system and driver models can completely or partially replace the active control of drivers [3, 4], which has a great potential in reducing traffic accidents and alleviating traffic congestion. [5]. For the military application, UGV can effectively replace soldiers to complete various task on the battle field including intelligence acquisition, surveillance and reconnaissance [6]. Thus, UGV technology can greatly promote the construction of intelligent transportation and the development of unmanned combat[7].

Environmental perception is one of the most important technologies in UGV, including the acquisition of external environmental information and the state estimation for the vehicle itself[8]. In addition, algorithms for environmental perception require sensors to work, a complete UGV needs to be equipped with various sensors to ensure its safe and stable operation. Thus, high precision and reliability sensors are crucial to complete the scheduled task.

In this paper, our main contribution is to present a brief review of various sensors as well as their application on existing platforms. The paper is organized as followed: section 2 introduces the commonly used sensors for UGV as well as compares their strengths and weaknesses; section 3 summarizes sensor applications in some existing platform; section 4 present some hotspots and development direction for sensor technologies.

## II. Sensors for UGV

This section presents some presentative sensors which are widely used in UGV as well as sensors with great application potential.

UGVs employ a wide selection of sensors, which can be divided into exteroceptive sensor and proprioceptive sensors. Exteroceptive sensors are required for external environmental perception system to acquire knowledge of the surroundings; while proprioceptive sensors are equipped for state estimation system to monitor platform status in real time[9].

### A. Exteroceptive Sensors

Exteroceptive sensors including Lidar, radar, ultrasonic, monocular camera, stereo camera, Omni-direction camera, infrared camera and event camera are presented in this section. Furthermore, a detailed comparison of the above sensors is given in TABLE I. .

#### 1) Lidar

*a) Brief Introduction*

Lidar is an active sensor which is able to detect the position and speed of the target by emitting laser beam; the direction of laser beam can be changed all the time by a rotating mirror driven by a high-speed electric motor, reaching a 360 degrees horizontal detection range. Lidar can be divided into 2D-Lidar (only 1 laser beam) and 3D-Lidar (usually 4 to 128 laser beams) according to the number of laser beams. Both can construct point cloud map of the surroundings with high accuracy.

*b) Application*

Lidar is a good choice for SLAM, object detection, object tracking and point cloud matching, some typical algorithms using Lidar are as follows.

**SLAM:** Classic Lidar-based SLAM methods include Gmapping [10], Hector [11], Karto [12] for 2D SLAM ; Loam [13] for 3D SLAM; Cartographer [14] for both 2D and 3D SLAM. All the above methods are open sourced in ROS (Robot Operating System). Recent state-or-art methods are established mainly based on deep learning. Sascha Wirges [15] trained a network to construct a multi-layer grid map; Ji [16] proposed robust CPLG-SLAM that is suitable for off-road environment. Shen [17] put forward a optimized with fully CNN and dynamic local NDT, which showed a higher efficiency than methods with conventional NDT.

**Object detection:** Algorithms about object detection can be mainly divided into three categories: projection method, voxel grid method and point cloud network method. Numbers of tasks have been carried out for object detection using Lidar,

TABLE I. INFORMATION OF DIFFERENT EXTEROCEPTIVE SENSORS

| Sensors \ Features | Affected by Illumination | Affected by Weather | Color and Texture | Depth | disguised | Range | Accuracy | Size | Cost |
|---|---|---|---|---|---|---|---|---|---|
| Lidar | — | √ | — | √ | Active | <200m | High | Large | High |
| Radar | — | — | — | √ | Active | <250m | Medium | Small | Medium |
| Ultrasonic | — | — | — | √ | Active | <5m | Low | Small | low |
| Monucular Camera | √ | √ | √ | — | Passive | — | High | Small | low |
| Stereo Camera | √ | √ | √ | √ | Passive | <100m | High | Medium | low |
| Omni-direction Camera | √ | √ | √ | — | Passive | — | High | Small | low |
| Infrared Camera | — | √ | — | — | Passive | — | Low | Small | low |
| Event Camera | √ | √ | — | — | Passive | — | Low | Small | low |

\* The range of cameras except for depth range of stereo camera is related to operation environmental thus there is no fixed detection distance

the principle, pros and cons, and some related works are shown in TABLE II. .

**Object tracking:** The large field of view and high precise depth information allow Lidar track the object over a long periods of time.

Most widely used technologies rely on data association, which means to match the target at the current moment with the tracking target of the previous time cycle according to a similarity judgment. Specific methods include nearest neighbor method [18], multiple hypothesis tracking [19], similarity metrics with point density [20] or Hausdorff distance [21], and Bernoulli filter [22].

Physical models for object tracking are also carried out by some researchers including dynamic model [23] and shape model [24].

In addition, the results of object tracking often require noise reduction to achieve a smoother tracking result. In general, filter methods are commonly used including Kalman filter [25] and particle filter [26].

**Point cloud matching:** Point cloud matching is usually used to find the area with high similarity between online point clouds and high-precision priori point cloud maps to achieve localization.

In general, there are two classic point cloud matching methods, including Iterative Closest Point (ICP) [27] and Normal Distribution Transform (NDT) [28]; the two methods were compared in [29], and NDT proved to be more robust than ICP. Several researches were proposed to improve the performance of the above two methods, specifically MCL [30], IMLP (optimize matching accuracy of ICP) [31] and adaptive-NDT (reduce noisy causing by NDT) [32].

*c) Pros and Cons*

The strength of Lidar is its long detection distance ranging, wide field of view and high accuracy of data collection. In addition, depth information of target can be obtained from Lidar, and it is not affected by the lighting conditions which means it is able to work at both day and night.

However, there are several weaknesses to be taken into account. First, it has extremely high cost. What's more, it suffers from sparse point cloud and low vertical resolution at a long distance, which may result in false detection. In addition, data collected by Lidar is easily affected by weather such as rainy, foggy, snowy and sandstorm [33]. Last but not least, it has poor performance in dark and specular object detection since they absorb or reflect most radiation of laser beams.

*2) Radar*

*a) Brief Introduction*

Radar plays a important role in both civil and military field. The principle of radar is similar to that of Lidar, but it emits radio waves instead of laser beam to detect the position, distance and velocity of the target. Radar can be divided into various types according to different bands, of which millimeter wave radar (MMW) is the most commonly used for UGV.

*b) Application*

Radar is mainly used for object detection, and ADAS system. Several researches are list as follows:

**Object detection:** Algorithms about object detection can be mainly divided into two categories: machine learning methods and end-to-end network methods.

Widely used machine learning method in radar detection are LSTM and random forest. Ole Schumann [34] clustered radar data based on the DBSCAN and compared the effectiveness of random forest and LSTM methods in vehicle detection. Tokihiko Akita [35] extracts candidate regions based on radar echo intensity, and uses LSTM to detect and track targets. End-to-end net work is similar to point-net method applied in Lidar, including improved PointNet [36] and PointNet++ [37].

**ADAS:** Radar-based ADAS related work mainly includes blind spot detection [38], lane change assistance [39] and collision warning [40], which is able to improve vehicle safety and assist drivers with their decision-making.

*c) Pros and Cons*

Radar sensors outperform for its high performance price ratio. Compared with Lidar, radar has a further detection range with small size and low cost, in addition it is not easily affected by light and weather condition. Apart from the above advantages, several weaknesses of radar are listed below: low resolution and accuracy, filters are usually needed to use for preprocessing [33]; limited for color and texture information; poor concealment and easy interference with other equipment [41].

TABLE II. REASEARCHES OF OBJECT DETECTION USING LIADR

| Methods | | Principle | Literature | Advantages | Limitations |
|---|---|---|---|---|---|
| Projection | Spherical | Project the point cloud to spherical coordinate system including azimuth, elevation, and distance. | • SqueezeSeg [42]; <br> • PointSeg [43] | • Point cloud get more dense after transformation | • Difficult to achieve sensor fusion |
| | Plane | Project the point cloud to main view to generate depth information. | • Faster R-CNN [44] <br> • ConvNet [45] | • Convince for data fusion with camera images | • Empty pixels may be produced at distant location due to sparse point cloud. |
| | Bird-Eye | Project the point cloud to top-view image in order to provide size. and position information | • Deep learning: PIXOR [46]; YOLO3D [47] <br> • Other: Similarity [48] | • Directly provide the location and size information of the object | • Sparse point cloud at distant location may cause error detection |
| Volumetric | | Assign the point cloud to the three-dimension voxel grid at the corresponding position. | • Deeplearning: Voxelnet [49]; Second [50]; MVX-Net [51] <br> • Other: Particle filter [52] | • Original 3D data information can be retained | • Empty voxel grids are generated due to sparse and uneven distribution of point cloud |
| PointNet | | Directly take the original point cloud as input and achieve end-to-end detection. | • PointNet [53] <br> • PointNet++ [54] <br> • PointPillars [55] | • Simple and fast <br> • No hard demand for point cloud pre-processing | • Usually a long network training period |

TABLE III. REASEARCHES OF OBJECT DETECTION USING MONOCULAR CAMERA

| Methods | | | Principle | Literature | Advantages | Limitations |
|---|---|---|---|---|---|---|
| Two-stage Detection | HG | Apperence-based | Prior knowledges of the object are required to generate ROI. | • Color [56] <br> • Edge [57] <br> • Texture [58] | • High accuracy <br> • Various type of features can be extracted | • High computing cost <br> • Easily affected by environment |
| | | Motion-based | Temporal information are extracted to generate ROI. | • Fame-difference [59] <br> • Bckgroud modeling [60] <br> • Otical flow [61] | • Low computing cost <br> • Easy to implement | • Hard for low speed and stationary objects <br> • Hard for complex scences |
| | HV | Template-based | Establish feature template and calculate the ximilarity measurement. | • Hybrid template [62] <br> • Deformable template [63] | • Low computing cost <br> • Simple algorithm | • Difficult to build a comprehensive template library |
| | | Learning-based | Train the classifier to verify the target in the candidate region. | • SVM <br> • Adaboost <br> • Neural network [53] | • High accuracy <br> • Robust | • High computing cost <br> • Hard to train a classfier |
| One-state Detection | | | Achieve end-to-end detection through deep learning with out HG process. | • SSD [64]; YOLOv3 [65]; MB-Net [66]; EZ-Net [67]; | • Fast detection speed and good real-time performance | • Low accuracy |

*3) Ultrasonic*

*a) Brief Introduction and Application*

Ultrasonic detects targets by emitting sound waves, whose principle is similar to Lidar and radar. Ultrasonic is widely used in the field of ships; however for UGV, it is capable to be equipped for detecting short-range target [68], ADAS system including automatic parking [69] and hazard warning [70].

*b) Pros and Cons*

Ultrasonic is not easily affected by light and weather condition, and has a low cost and small size. However, the detection distance of ultrasonic is short, the accuracy is low which is easy to generate noisy in collected data [71]. In addition, the narrow field of view usually causes blind spots during detection work.

*4) Monocular camera*

*a) Brief Introduction*

Monocular camera stores environmental information in the form of pixels by converting optical signals into electrical signals. The image collected by the monocular camera is basically the same as the surroundings sensed by human eye. Monocular camera is one of the most popular sensors in UGV fields, which is strongly capable to many kinds of tasks for environmental perception.

*b) Application*

A numbers of researches have been carried out by using monocular camera: SLAM, semantic segmentation, object detection, road detection and traffic sign detection.

**SLAM:** SLAM based on monocular camera are commonly used for environmental perception due to its low cost and convince, however it can't construct a true size map due to the lack of depth information. Classic monocular camera based SLAM methods include MonoSLAM [72], ORB-SLAM [73], ORB-SLAM v2 [74] and DTAM [75]; and recent algorithms include LSD-SLAM [76], SVO [77] and InertialNet [78].

**Semantic segmentation:** Semantic segmentation is essential for the understanding of the driving environment and the extraction of objects of interest. Conventional algorithms include Conditional Random Fields (CRF) [79] and Markov Random Fields (MRF) [80] while recent methods are mainly developed basing on deep learning to improve segmentation accuracy, specifically FCN [81], PSPNet [82], DeepLab [83], ChiNet [84] and SSeg-LSTM [85].

**Object detection:** Numerous works have been carried out for object detection using monocular camera, which can be divided into two-stage detection method and single-state detection method. For two-stage detection method, the first stage called hypothesis generation (HG) generates region of interests (ROI) of the object, the second stage called hypothesis verification (HV) trains a classifier to identify the candidate regions; while single-stage method relies on the neural networks to achieve an end-to-end detection. Several related works are shown in TABLE III. .

TABLE IV. REASEARCHES OF OBJECT DETECTION USING STEREO CAMERA

| Methods | Principle | Literature | Advantages | Limitations |
|---|---|---|---|---|
| IPM | Transform data from the camera coordinate system to the world coordinate system to build top-view image without disparity. | [86] | • Low computing cost<br>• Simple and mature algorithm | • Vulnerable to road conditions including off-road and uneven road |
| Disparity Map | Extract plane features parallel (perpendicular) to the camera plane which may may contain object by generating a disparity map | • V-disparity map [87]<br>• UV-disparity map [88]<br>• Filter [89] | • High accuracy<br>• Easy to obtain depth information | • High computing cost<br>• Low resolution for planes with similar shapes |

TABLE V. REASEARCHES OF ROAD DETECTION USING MONOCULAR CAMERA

| Methods | | Principle | Literature | Advantages | Limitations |
|---|---|---|---|---|---|
| Feature Extraction | Apperence-based | Detect road area by means of road appearance. | • Color[90]<br>• Texture [91] | • Different type of features can be used | • Difficult to detect unstructed road |
| | Boundary-based | Detect elevation difference between the road and its surrounding to generate road boundary. | [92] | • Suitable for most scenarios | • Difficult to detect off-road with unclear boundary |
| Model Fitting | Parametric | Fitting the road model with parametric equations. | • Line [93]<br>• Polynomial [94] | • Low computing cost<br>• Simple model | • Low accuracy<br>• Poor robustness |
| | Semi-parametric | Fitting the road model with semi-parametric without prior assumtion of road model. | • Cubic-spline [95]<br>• B-spline [96] | • No need to assume a specific geometry of the road | • High computing cost<br>• Over-fitting need to be considered |
| | Non-parametric | Fitting the road model with a curve only requires continuity. | • Hierarchical-Bayesian [97]<br>• Dijkstra [98] | • Only continuity of the curve need to be required | • High computing cost<br>• Complicate model |

**Road detection:** Road detection is an important work to make sure that platform can be stably driven in a safe and accessible area of the road. Algorithms can be mainly divided into two stages: feature extraction and model fitting. Several related works are shown in TABLE V. .

**Traffic sign detection:** Correct identification of traffic signs is essential to the platform's reasonable behavior decision. In general, detection of traffic sign can be divided into two steps: segmentation and verification.

The purpose of segmentation is to obtain the location of the traffic sign; color-based method including RGB [99], HSV [100] and HIS [101] color space, and shape-based method including edge [102], HOG [103] and Haar wavelet feature[104] are often used for segmentation. The purpose of verification is to identify the specific type of traffic sign from the result of segmentation, a serious of learning-based method are mainly carried out specifically SVM [103], AdaBoost [105] and neural network [106].

*c) Pros and Cons*

The biggest advantage of monocular camera is that it can generate high-resolution images with color and texture information from the environment; as a passive sensor, it also has good concealment and will not interfere with other devices. In addition, camera-based perception technology is relatively mature, and the size of monocular camera is small with low-cost. Nevertheless, several drawbacks of monocular camera should also be considered: lack of depth information; highly affected by illumination and weather condition; large calculation cost for high-resolution images.

5) *Stereo camera*
   *a) Brief Introduction*

The imaging principle of the stereo camera is the same as that of the monocular camera, but the stereo camera is equipped with an additional lenses at a symmetrical position or calculating the time of flight to generate depth information; another solution is to place two or more monocular cameras at different positions on the platform to form a stereo vision system, however this will bring greater difficulty to the calibration of the camera.

*b) Application*

Stereo camera is widely used in the following work: SLAM, object detection and road detection.

**SLAM:** Depth information collected by stereo camera make it more suitable to generate precise 3D maps than monocular camera. Relevant methods include DVO [107], RGBD-SLAM-V2 [108] and Stereo-ORB-SLAM [109].

**Object detection:** Algorithms developed within stereo camera are similar to that of monocular camera. Since stereo camera is able to collect depth information, it has more choices in the ROI generation step including inverse perspective mapping (IPM) and disparity map. Several related work are shown in TABLE IV. .

**Road detection:** Road detection methods using stereo camera are similar to that of monocular camera. Inverse perspective mapping (IPM) [110] are usually carried out to achieve road detection since the bird-eye's view construct from IPM is able to intuitively extract geometric information of the road, thereby facilitating the extraction of road features and model fitting.

TABLE VI. SENSOR APPLICATION IN SOME EXISTING UGV

| Basic Information | | | Sensor Application | | Platform Perfromance |
|---|---|---|---|---|---|
| Name | Time | Country | *Exteroceptive* | *Proprioceptive* | |
| CITAVT-IV [111] | 2002 | China | 2 monocular cameras for object and road detection | Odometer and IMU for localization | Highest speed 75.6km/h on Changsha Around-city Express |
| THMR-V [112] | 2002 | China | 1 Lidar and 2 monocular cameras and for object and road detection | GPS for localization | Average speed 5~7km/h in Tsinghua University |
| Junior [113] | 2007 | America | 5 Lidars for SLAM; 1 monocular camera for road detection; 2 radars for object detection | GPS and IMU for localization | Average speed 20km/h in urban environment of 2007 Darpa Challenge |
| Boss [114] | 2007 | America | 9 Lidars for SLAM and object detection; 2 radars for object detection; 2 monocular cameras for road detection | Integrated navigation for localization and state estimation | Average speed 22.53km/h in urban environment of 2007 Darpa Challenge |
| Odin [115] | 2007 | America | 8 Lidars and 2 monocular cameras and for object and road detection | GPS and IMU for localization; IMU for state estimation | Average speed 20.92km/h in urban environment of 2007 Darpa Challenge |
| Talos [116] | 2007 | America | 13 Lidars and 15 radars for object detection; 5 monocular cameras for road detection | GPS and IMU for localization; IMU for state estimation | Urban environment of 2007 Darpa Challenge. (speed not mentioned) |
| Robotcar [117] | 2013 | U.K | 3 Lidars, 2radars, 3 monocular cameras and 1 stereo camera for vehicle, pedestrain and road detection | GPS and IMU for localization; IMU for state estimation | Operated more than 100 times around Oxford city centre. (speed not mentioned) |
| BRAiVE [118] | 2013 | Italy | 5 Lidars for object detection; 1 radar for ADAS; 10 monocular cameras for road and traffic sign deetection | GPS and IMU for localization; IMU for state estimation | 13,000km section from Italy to Shanghai. (speed not mentioned) |
| Bertha [119] | 2014 | Germany | 6 radar for object detection; 2 monocular cameras for traffic sign deetection; 1 stereo camera for road detection. | GPS for localization | Average speed 30km/h from Mannheim to Pforzheim (103km). |
| Uber | 2015 | America | 1 Lidar, 10 radars and 7 cameras for SLAM, object and road detection | GPS and IMU for localization; IMU for state estimation | Have been tested in city section of San Francisco, Pennsylvania. |
| Land Cruiser [120] | 2016 | China | 1 Lidar for SLAM; 4 monocular cameras for object detection | GPS and INS for localization; | Have been tested in off-road environment.. |
| Autopilot | 2017 | America | 1 radar and 10 ultrasonics for ADAS; 5 monocular cameras for road detection. | GPS for localization | Not mentioned |
| Apollo [121] | 2018 | China | 1 Lidar, 1 radar, 1 ultrasonic and 6 monocular cameras for ADAS, object and road detection, traffic sign detection | GPS and IMU for localization; IMU for state estimation | Good perfromance in the Chinese Subject Three Examination |

*c) Pros and Cons*

Compared with Lidar, stereo camera can obtain color information and generate denser point clouds [122]; compared with monocular camera, two images taken simultaneously through multiple angles of view by different lens can obtain depth information and motion information of the environment.

However, stereo camera is also susceptible to weather and illumination conditions; in addition, the field of view is narrow; and additional calculation is required to process extra depth information [122].

*6) Omni-direction camera*

*a) Brief Introduction and Application*

Compared with monocular camera, an Omni-direction camera is able to collect a ring-shaped panoramic image centered on the camera. With the development of hardware, it is gradually used in UGV including SLAM [123] and semantic segmentation [124].

*b) Pros and Cons*

The Omni-direction camera has a large field of view, however the computational cost is large due to the increase collection of image point clouds.

*7) Infrared camera*

*a) Brief Introduction*

Infrared camera gathers surroundings information by receiving signals of infrared radiation from objects, which is able to complement well with conventional cameras. In general, infrared camera can be divided into the active infrared camera operating in the near-infrared (NIR) region and passive infrared camera operating in the far-infrared (FIR) region. The NIR camera is sensitive to the wave lengths in the range of 0.15-1.4 μm; while FIR camera is sensitive to the wave lengths in the range of 6-15 μm.

*b) Application*

Infrared camera is popular for object detection (usually vehicles and pedestrian) at night for UGV, and it should be noticed that the corresponding type of camera should be selected according to the infrared wavelength range emitted by the detected object. Several researches among infrared camera have been put forward as followed:

Some researchers established the method based on appearance or feature extracted from the image, specifically edge [125], HOG [126] and Haar [127]. However the low resolution of infrared images makes it difficult to extract features; it is a good choice to enhance the contrast of infrared images first, so that the subsequent detection work can achieve better results [127]. Other relevant works were mainly based on YOLOv3 [128] network in deep learning.

Construct a stereo infrared vision system is another way to solve the detection problem. Mita [129] placed two infrared cameras to form a stereo infrared vision system, and achieve vehicle detection under different weather conditions by calculating disparity map.

*c) Pros and Cons*

The biggest advantage of the infrared camera is that it has good performance at night; and it is low-cost, small in size. However, it cannot generate color, texture, and depth information, and the resolution of the image is relatively low.

*8) Event camera*

*a) Brief Introduction*

Event camera is a newer perception modalities applied in the field of UGV. The working principle of event camera has a large difference comparing with traditional cameras, it records a series of asynchronous signals by measuring the change in brightness of each individual pixel of the image at the microsecond level [130].

*b) Application*

Event camera is suitable for highly dynamic application scenarios of UGV such as SLAM [131], state estimation [132] and object tracking [133].

*c) Pros and Cons*

The recorded events are sparse in both space and time, which has a large potential to reduce the time of information transmission and processing. In addition, the microsecond level time resolution makes it maintain a high dynamic measurement range. However, the resolution of output data is low.

*B. Proprioceptive sensors*

proprioceptive sensors including GNSS and IMU are summarized in this section.

*1) GNSS*

Global Navigation Satellite System (GNSS) is the most widely used technology for UGV localization system. The current mainstream GNSS systems are GPS developed by United States, GALILEO developed by European Union, GLONASS developed by Russia, and Beidou System developed by China. It is able to provide absolute position information for UGV through a serious of satellites around the earth.

The working principle of GNSS is based on measuring the time of flight of the signals transmitted by satellites and received by receptor. Position coordinate and pseudo-range can be obtained in the form of $(x, y, z, t)$, where x, y and z represent coordinates while t represents pseudo-range.

There are several factors that affect the positioning accuracy of GNSS: errors caused by satellites, including the atomic clocks and orbit error; errors caused by signal transmission, including transmission in ionosphere and troposphere and multipath effect; errors caused by receptor noisy.

*2) IMU*

IMU stands for Inertial Measurement Unit, which includes accelerometer, gyroscope and magnetometer. It is mainly used to measure the acceleration and angular velocity information of the UGV. It allows to monitor the running status of the vehicle and provide real-time feedback for the bottom control of the platform. IMU often performs data fusion positioning with GPS to improve its positioning accuracy.

## III. APPLICATION OF SENSORS IN EXISTING PLATFORMS

This section introduces some existing UGVs, All the platforms are arranged in TABLE VI. , with the focus on sensor application of each platforms as well as their basic information and test performance.

## IV. FUTURE OUTLOOK

The related hotspots around sensor technologies are as follows.

*A. Sensor Fusion*

Usually, there are two mainly problems need to be solved for sensor fusion: what sensors should be fused; how to assign the work of different sensors. In any case, the advantages of different sensors must be complemented in order to achieve a higher accuracy and stronger robustness system. However, the real-time performance and sensor cost need to be weighed since the equipment of multiple sensors will increase the complexity of the model and algorithm.

*B. Unusual Object Detection*

In addition to the efficient detection of conventional objects such as vehicles, pedestrians and buildings, objects in some special scenarios including negative obstacles, objects in a very short or very long distance also need to be correctly identified.

Negative obstacles including potholes, holes and trenches can threaten driving safety of UGV; objects in a very short distance such as pedestrian beside the vehicle are often in blind spots of sensors while objects in a very long distance may cause error identification because of low resolution. Generally speaking, unusual obstacle detection is a great guarantee for the safety of the platform, more researches need to be carried out in the future.

*C. Application in High-speed Platform*

The high-speed driving of UGV is an important guarantee for its maneuverability, which is essential to complete the task efficiently. However, if the sensor algorithm cannot generate surrounding information at high speed, the platform is extremely prone to danger during high-speed driving. Thus, how to effectively improve the real-time performance of the sensor technologies and validate the effect through substantial vehicle test need to be highly considered in the future.

## V. CONCLUSION

Sensor technologies play a more and more important role in the development of UGV. This paper proposes a brief review on sensor technologies for UGV, including their brief introduction, algorithms, strengths and weakness, as well as the application in some existing UGVs. Finally, the hotspots of sensor technologies are forecasted to highlight the research direction.

In general, there are still many aspects that need to be improved around sensors, which is of great research value. Future work should focus on the survey of sensor fusion technologies as well as sensor applications in future developed platforms.